\documentclass[conference]{IEEEtran}
\IEEEoverridecommandlockouts
\usepackage{cite}
\usepackage{amsmath,amssymb,amsfonts}
\usepackage{algorithmic}
\usepackage{graphicx}
\usepackage{CJK}
\usepackage{textcomp}
\usepackage{xcolor}
\usepackage{bm} 
\usepackage[hyphens]{url} 
\usepackage{algorithm}
\usepackage{algorithmic}
\usepackage{booktabs}
\usepackage{multirow}
\usepackage{multicol}

\def\BibTeX{{\rm B\kern-.05em{\sc i\kern-.025em b}\kern-.08em
    T\kern-.1667em\lower.7ex\hbox{E}\kern-.125emX}}
\begin{document}

\title{Compose Like Humans: Jointly Improving the Coherence and Novelty for Modern Chinese Poetry Generation}
%\title{Conference Paper Title*\\

\def\first{$^1$}
\def\second{$^2$}
\def\third{$^3$}
\def\comma{$^,$}
\def\star{$^*$}

\author{
\IEEEauthorblockN{Lei Shen\first\comma\second, Xiaoyu Guo\third, Meng Chen\third\star}
\IEEEauthorblockA{\first\textit{Key Laboratory of Intelligent Information Processing}, \\ \textit{Institute of Computing Technology, Chinese Academy of Sciences}, Beijing, China}
\IEEEauthorblockA{\second\textit{University of Chinese Academy of Sciences}, Beijing, China}
\IEEEauthorblockA{\third\textit{JD AI}, Beijing, China}
shenlei17z@ict.ac.cn, guoxiaoyu404@163.com, chenmeng20@jd.com
}

\iffalse
\author{\IEEEauthorblockN{Lei Shen}
\IEEEauthorblockA{\textit{Key Laboratory of Intelligent Information} \\
\textit{Processing, Institute of Computing Technology,}\\
\textit{Chinese Academy of Sciences,} \\
Beijing, China \\
shenlei17z@ict.ac.cn}
\and
\IEEEauthorblockN{Xiaoyu Guo}
\IEEEauthorblockA{\textit{JD AI,} \\
Beijing, China \\
guoxiaoyu404@163.com}
\and
\IEEEauthorblockN{Meng Chen}
\IEEEauthorblockA{\textit{JD AI,} \\
Beijing, China \\
chenmeng20@jd.com}
}
\fi

\maketitle

\newcommand\blfootnote[1]{% 
\begingroup 
\renewcommand\thefootnote{}\footnote{#1}% 
\addtocounter{footnote}{-1}% 
\endgroup
}

\begin{abstract}
\blfootnote{\star{Corresponding Author.}}Chinese poetry is an important part of worldwide culture, and classical and modern sub-branches are quite different. The former is a unique genre and has strict constraints, while the latter is very flexible in length, optional to have rhymes, and similar to modern poetry in other languages. Thus, it requires more to control the coherence and improve the novelty. In this paper, we propose a generate-retrieve-then-refine paradigm to jointly improve the coherence and novelty. In the first stage, a draft is generated given keywords (i.e., topics) only. The second stage produces a ``refining vector'' from retrieval lines. At last, we take into consideration both the draft and the ``refining vector'' to generate a new poem. The draft provides future sentence-level information for a line to be generated. Meanwhile, the ``refining vector'' points out the direction of refinement based on impressive words detection mechanism which can learn good patterns from references and then create new ones via insertion operation. Experimental results on a collected large-scale modern Chinese poetry dataset show that our proposed approach can not only generate more coherent poems, but also improve the diversity and novelty.
\end{abstract}

\begin{IEEEkeywords}
generate-retrieve-then-refine paradigm, automatic poetry generation, coherence and novelty
\end{IEEEkeywords}

\section{Introduction}
Automatic poetry generation is a sub-field of Natural Language Generation (NLG). In recent years, there have been many studies focusing on the classical Chinese poetry generation, since this kind of poetry is distinctive. Among different types of classical poems, {\it quatrain} (\begin{CJK*}{UTF8}{gbsn}绝句\end{CJK*}) and {\it regulated verse} (\begin{CJK*}{UTF8}{gbsn}律诗\end{CJK*}) are perhaps the best-known ones. They mainly have four requirements: 1) strict constrains in length, e.g., a {\it quatrain} consists of four lines, and each line contains five or seven characters; 2) tonal patterns, i.e., ``{\it Ping}'' (level tone) or ``{\it Ze}'' (downward tone); 3) rhyme schemes, e.g., for a {\it quatrain}, the ending character of the second and the fourth lines should have the same rhyme; 4) unified structure, e.g., a {\it quatrain} often follows the ``beginning, continuation, transition, summary'' template \cite{he2012generating}.

\begin{figure}[htb]
\begin{center}
   \includegraphics[width=1.0\linewidth]{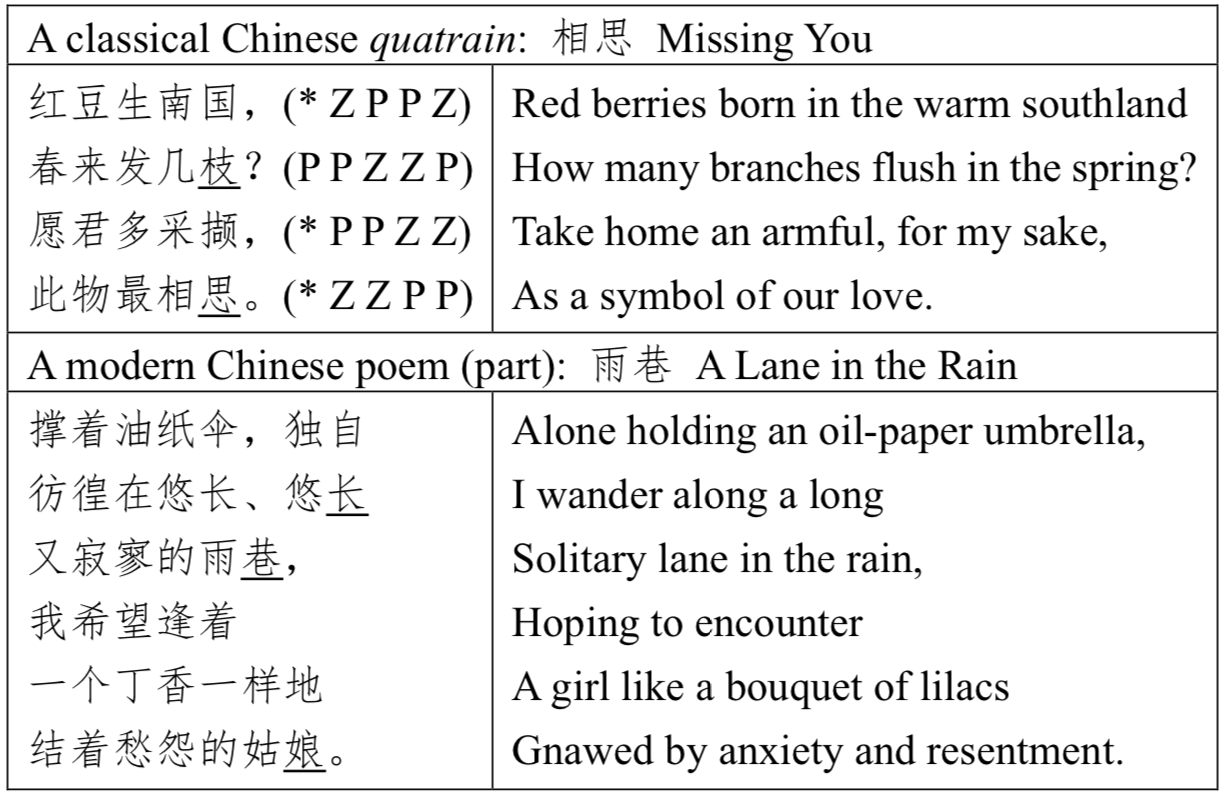}
\end{center}
   \caption{A comparison between the classical and modern Chinese poetry. The upper one is a 5-char {\it quatrain} exhibiting one of the most popular tonal patterns, which is also used in Zhang and Lapata's paper \cite{zhang2014chinese}. The tone of each character is shown within parentheses, P and Z are ``{\it Ping}'' and ``{\it Ze}'', respectively. * indicates that the tone is not fixed and can be either. The lower one is part of a famous modern Chinese poem. Rhyming characters are shown with underlines.}
\label{fig:example}
\end{figure}

The modern Chinese poetry has become more and more popular nowadays, and people use it to record daily life, express personal emotions, and send blessings at special occasions. It is similar to modern poetry in other languages, and does not have too many strict constraints. Meanwhile, there are some challenges for automatic modern Chinese poetry generation. Linguistic accordance (coherence) and aesthetic innovation (novelty) are two important aspects. Modern poems are more free in length, thus it is hard to control the coherence. Besides, writing poems is an artistic creation process so novelty is necessary, which means more imagination and various uses of language are needed \cite{cheng2018image}. However, recent works mainly focused on the classical Chinese poetry and could not cover both two aspects very well at the same time \cite{zhang2017flexible}.

To improve the coherence and novelty simultaneously, we borrow thoughts from how humans compose a poem. Not like one-stage automatic poetry generation, humans tend to start with a draft, and keep polishing it. Basically, there are two ways to refine a draft: 1) learning from predecessors' works. They learn how to use words and organize sentences from others' masterpieces, and then apply them into their works to create new expressions; 2) deliberating one sentence based on the context. With the information from previous and following sentences, they can modify current sentence to fit in the whole poem more appropriately. There have been some works imitating above ideas individually, and Wu et al. \cite{wu2018response} summarized them as either ``retrieve-then-generate'' paradigm \cite{yan2016poet,xia2017deliberation,wang2018paper} or ``generate-then-refine'' paradigm \cite{song2016two,pandey2018exemplar,cao2018retrieve,li2018delete,guu2018generating,wu2018response,shen2019modeling}. However, we think both two ways are necessary and need to be considered together, so we bring up the ``generate-retrieve-then-refine'' paradigm. Besides, previous works can only utilize history sentences or word-level bidirectional context, or they simply feed all retrieval lines into model which may contain lots of noises.

To tackle the above problems, we propose a novel approach that polishes generated drafts with bidirectional sentence-level context and a ``refining vector'' for modern Chinese poetry generation. In the first stage, the model generates a draft which provides future sentence-level information. Second, it leverages the generated draft to get some retrieval lines, and uses the impressive words detection mechanism to get the ``refining vector''. At last, both bidirectional sentence-level context and the ``refining vector'' are applied to generate a refined poem. %Previous works can only utilize history sentences or word-level bidirectional context, or they simply feed all retrieval lines into model which may contain lots of noises. 
Since we use both the past and future information in sentence level, we can improve the coherence of the entire poem better. By using impressive words detection mechanism, we filter out noises and extract some good expression patterns to distill the ``refining vector'', and finally improve the novelty and diversity of language usage.

Since there is no public large-scale modern Chinese poetry dataset\footnote{Liu et al. \cite{liu2019rhetorically} published a small modern Chinese poetry dataset with 60,000 sentences in total. Their poems are cut into short chunks with the size of 3 lines, while our dataset has over 9 million lines and keeps the original long poems.}, we collect one from the internet, and will publish it in the near future. Experimental results on this dataset show that our approach outperforms other baselines significantly in terms of the coherence and novelty.

%Our contributions are listed as follows: 1) this paper proposes a new paradigm, generate-retrieve-then-refine, for poetry generation; 2) we leverage future information from the draft and the ``refining vector'' produced by impressive words detection mechanism ; 3) we collect a large-scale modern Chinese poetry dataset, and empirically verify the effectiveness of our model to jointly improve the coherence and novelty of generated poems.
%\begin{itemize}
%\item Bidirectional sentence-level context is used to achiever better coherence.
%\item We apply a ``refining vector'' produced by impressive words detection and insertion mechanism to enhance the novelty of drafts.
%\item When generating one line in second stage, we leverage bi-directional context provided by the draft.
%\item We utilize impressive words detection and insertion mechanism to learn good expression patterns from retrieval results.
%\item We collect a large-scale modern Chinese poetry dataset, and empirically verify the effectiveness of our model to jointly improve the novelty and diversity on it. 
%\end{itemize}

Our main contributions can be summarized as follows: 
\begin{itemize}
\item We propose a new paradigm, generate-retrieve-then-refine, for automatic poetry generation.
\item In order to jointly improve the coherence and novelty, we leverage future information from the draft and the ``refining vector'' produced by impressive words detection mechanism.
\item We collect a large-scale modern Chinese poetry corpus, and empirically verify the effectiveness of our model in terms of fluency, coherence, novelty and diversity.
\end{itemize}

%The remaining of the paper is organized as follows. Related work is first presented in Section \ref{sec:relatedwork}. Then, we introduce the research background in Section \ref{sec:background} and describe our approach in Section \ref{sec:approach}. Section \ref{sec:experiments} details our experimental settings and results. Finally, Section \ref{sec:conclusion} concludes the paper.

\begin{figure*}[!t]
    \centering
    \includegraphics[scale=0.43]{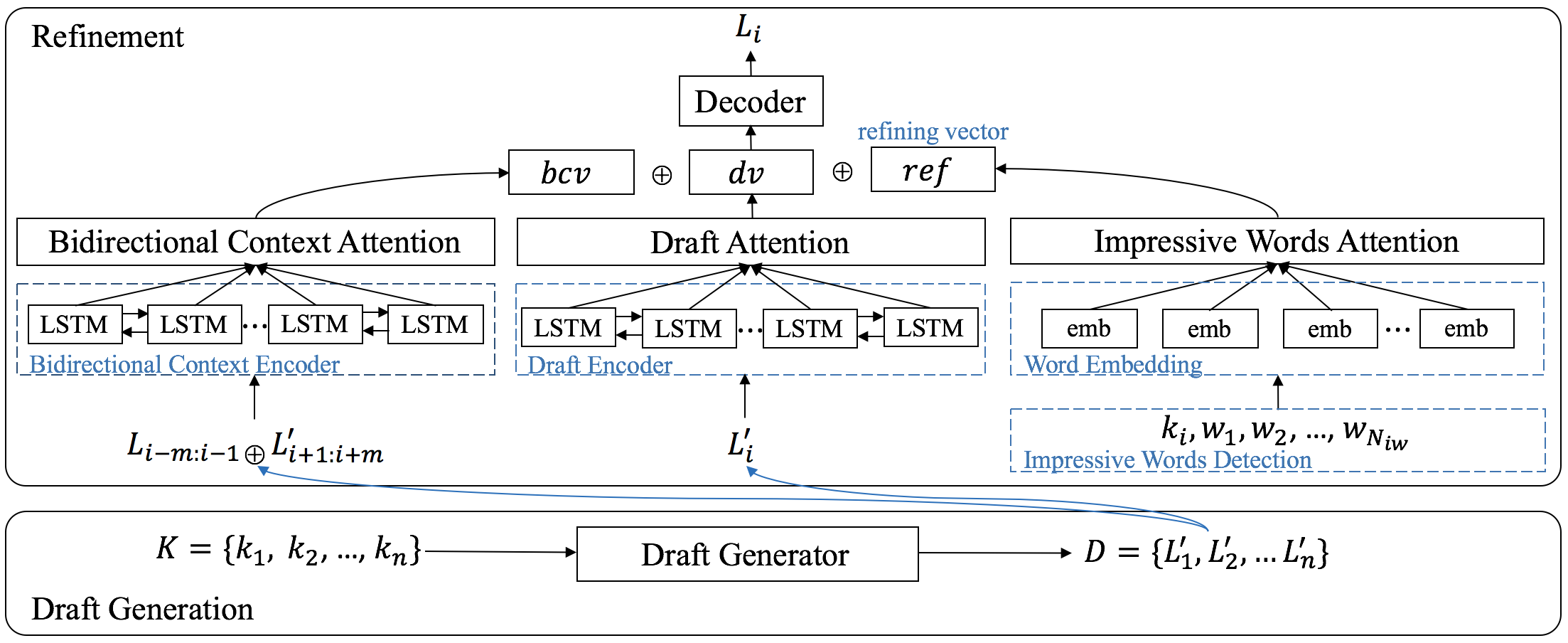}
    \caption{Overview of our GRR (generate-retrieve-then-refine) model. Lower: The draft generator generates draft $D$ given $n$ keywords. Upper: When generating line $L_i$, bidirectional context encoder is used to encode context ${L_{i-m:i-1}}$ and ${L}'_{i+1:i+m}$, draft encoder is for $L'_i$. Impressive word candidates consist of keyword $k_i$ and good patterns we detected. ``$bcv$'', ``$dv$'' and ``$ref$'' are bidirectional context vector, draft vector and refining vector, respectively.}
    \label{fig:model}
\end{figure*}

\section{Related Work}
\label{sec:relatedwork}
Our work touches two research fields: automatic poetry generation and refinement methods.
\subsection{Automatic Poetry Generation} 
Early researches in this area are based on grammatical rules \cite{oliveira2012poetryme,wu2009new,tosa2008hitch}, genetic algorithms \cite{manurung2012using,zhou2010genetic,manurung2004evolutionary} or statistical machine translation methods \cite{he2012generating,greene2010automatic,jiang2008generating}. After the boom of deep learning, many new approaches have appeared. Yan et al. \cite{yan2013poet} formulate this task as an optimization problem based on a generative summarization framework. Zhang and Lapata \cite{zhang2014chinese} utilize Recurrent Neural Networks (RNN) to take into account the entire history of what has been generated. Wang et al. \cite{wang2016chinese} propose a two-step method which first plans the sub-topics of the poem and then generates it with a modified Seq2Seq model. Yang et al. \cite{yang2017generating} employ a conditional variational autoencoder to generate thematic poems. Zhang et al. \cite{zhang2017flexible} leverage external memories to improve the creativity of generated poems. In order to achieve better coherence, Yi et al. \cite{yi2018chinese} propose a novel Working Memory model to keep a coherent information flow and learn to express each topic flexibly and naturally. Cheng et al. \cite{cheng2018image} generate Modern Chinese poetry from images. Liu et al. \cite{liu2019rhetorically} work on rhetorical patterns (e.g., metaphor and personification) in modern Chinese poetry.

Our work differs from the above since: 1) most of them are based on classical Chinese poetry generation; 2) the inputs are not the same, and our input is text but not images; 3) the points of interest are diverse, others focus on the theme, fluency, diversity, rhetorical patterns, etc., while we try to improve the coherence and novelty simultaneously.

%\subsection{Refinement Methods}
%The related work can be summarized into two categories: 1) generate-then-refine~\cite{yan2016poet,xia2017deliberation,wang2018paper}, a generative method with two or more iterations; 2) retrieve-then-generate~\cite{song2016two,pandey2018exemplar,cao2018retrieve,li2018delete,guu2018generating,wu2018response}, a combination of retrieval and generative methods. 

\subsection{Refinement Methods} 
There are two main paradigms for refinement. One is the ``retrieve-then-generate'' paradigm. Song et al. \cite{song2016two} and Pandey et al. \cite{pandey2018exemplar} encode all retrieval candidates into vectors and feed them into a decoder for response generation. Cao et al. \cite{cao2018retrieve} apply this paradigm in summarization by reranking and rewriting jointly. Li et al. \cite{li2018delete} similarly use deletion, retrieval and generation for text style transfer. Guu et al. \cite{guu2018generating} leverage latent variables to form the ``edit vector'' according to lexical differences (insertion and deletion words) in two sentences, while Wu et al. \cite{wu2018response} transfer the concept of ``edit vector'' to response generation and explicitly utilize the lexical differences in queries. The other is the ``generate-then-refine'' paradigm. Yan et al. \cite{yan2016poet} generate a quatrain based on several iterations. Xia et al. \cite{xia2017deliberation} propose Deliberate Network that uses one decoder to generate a prototype from scratch and another decoder to revise the prototype in a joint training way. When generating a word in a sentence, this model can leverage backward and forward words. Wang et al. \cite{wang2018paper} apply the Deliberation Network to abstract generation.  

The differences between the above paradigms and ours are listed below. ``Generate-then-refine'' paradigm only utilizes context information in word level, which means that when generating a word, it only looks backward and forward in the range of current sentence. In contrast, we use future sentences from the draft when generating a word, so we can see much wider context in sentence level. ``Retrieve-then-generate'' paradigm tries to edit the retrieval sentences. It cannot guarantee the content coherence as the retrieval sentences may be noisy and very different in topic and style. On the contrary, we try to edit a generated draft, as it is consistent to some extent, and we tune it and make it better. Besides, we distill a ``refining vector'' to point out the direction for refinement. The ``edit vector'' in previous works \cite{guu2018generating,wu2018response} is simply the concatenation of insertion and deletion words, while the ``refining vector'' represents good expression patterns in retrieval sentences, thus it contains more diverse language usage. Our approach is a combination of the above two paradigms.

\section{Background}
\label{sec:background}
%\subsection{Seq2Seq with Attention Mechanism}
For input $X = {\{x_i\}_{i=1}^{n}}$, where $n$ is the number of words, it is encoded into a sequence of hidden states $H = {\{h_i\}_{i=1}^{n}}$. Here we employ $\hat{e}_{x_t}$ as the embedding vector of word $x_t$, and the hidden state $h_t$ is defined as: 
\begin{equation} \label{eq: encoder}
    h_t = f_{\mathrm{LSTM}}(h_{t-1}, \hat{e}_{x_t}),
\end{equation}
where $f_{\mathrm{LSTM}}$ is the activation function of LSTM.

The decoder state $s_t$ is updated by:
\begin{equation} \label{eq:decoder_hidden_state}
    s_t = f_{\mathrm{LSTM}}(s_{t-1}, \hat{e}_{y_{t-1}}, c_t),
\end{equation} 
where $s_{t-1}$ and $\hat{e}_{y_{t-1}}$ are hidden state and word embedding of decoded word at time step $t-1$ respectively. $c_t$ is calculated by attention mechanism. Attention mechanism \cite{bahdanau2014neural} is designed to focus on input information which is highly related to the generation of current word. The relevance between the to-be-generated word $y_t$ and the $i$-th input word is computed as:
\begin{equation} \label{enegy function}
    r_{i,t}={\bm{\mathrm{v}}}_\alpha^T
    \tanh\left({\bm{\mathrm{W}}}_\alpha s_{t-1} + {\bm{\mathrm{U}}}_\alpha h_i\right).
\end{equation}
Then, the relevance score is normalized and serves as the weight for  corresponding encoder hidden state when calculating vector $c_t$:
\begin{equation}
    \label{eq:att_weigth_sum}
    \alpha_{i,t} = \frac{\exp \left( r_{i,t} \right)}{\sum_{i'=1}^{T} \exp \left( r_{i',t} \right)},
    c_t = \sum\nolimits_{i=1}^{n}\alpha_{i,t}h_i.
\end{equation}

\section{Approach}
\label{sec:approach}
Coherence of a poem is mainly embodied in the relevance between several consecutive sentences, while novelty can be regarded as the way how new expression patterns are constructed. We generate a draft given some keywords at first. Then, we improve coherence and novelty jointly by leveraging bidirectional sentence-level context and a ``refining vector'' from the impressive words detection mechanism. %Figure \ref{fig:overview} shows the overview of our proposed model.

\iffalse
\begin{figure}[!t]
    \centering
    \includegraphics[scale=0.55]{figs/overview.png}
    \caption{Overview of our generate-retrieve-then-refine approach. Given several keywords, a draft is generated at first. Based on the draft, we retrieve some candidate lines, and then the impressive words detection mechanism use them to produce a ``refining vector''. At last, previous generated lines, draft lines and the ``refining vector'' work together to generate a new poem.}
    \label{fig:overview}
\end{figure}
\fi

\subsection{Writing Topic Representation}
Suppose we have a dataset $\mathcal{D} = \{\hat{P}_j\}_{j=1}^{N}$, and $N$ is the number of poems. Each poem has $n$ lines, i.e. ${\hat{P}_j} = \{\hat{L}_i\}_{i=1}^{n}$. Following the works of Wang et al. \cite{wang2016chinese} and Yang et al. \cite{yang2017generating} which assume that each line in the poem corresponds to a keyword (sub-topic), we use TextRank \cite{mihalcea2004textrank} to extract keyword ${k_i}$ for each line. Then, we obtain $\{(\hat{L}_i, k_i)\}_{i=1}^{n}$ pairs for each poem.

TextRank is a graph-based algorithm. Each vertex stands for a candidate word and edges between two words represent their co-occurrence. Besides, the edge weight is set according to the total co-occurrence rate between these two words. The TextRank score $T(V_i)$ is initialized to a default value and computed iteratively until convergence according to the following equation:
\begin{equation}
\resizebox{.89\linewidth}{!}{$
    \displaystyle
    T(V_i) = (1-d) + d\sum_{V_j \in E(V_i)}\frac{w_{ji}}{\sum_{V_k \in E(V_j)} w_{jk}} T(V_j),
$}
\end{equation}
where $w_{ji}$ is the weight of the edge between vertex $V_j$ and $V_i$, $E(V_i)$ is the set of vertices connected with $V_i$, and $d$ is a damping factor. Empirically, $d$ is set to 0.85 and the initial score of $T(V_i)$ is 1.0.

\subsection{Draft Generation}
In Draft Generation Stage, our goal is to write a draft $D$ with $n$ lines, i.e. $D = {\{L'_i\}_{i=1}^{n}}$, given keywords ${\{k_i\}_{i=1}^{n}}$. $L'_i$ is generated by taking the concatenated result of keyword $k_i$ and previous $m$ lines $L'_{i-m:i-1}$ as input. We use a multi-layer encoder with bidirectional Long Short-Term Memory (LSTM) \cite{hochreiter1997long} to encode the input by concatenating the last hidden states of the forward and backward LSTMs of the top layer, i.e. $h_i = \left[{\overrightarrow{h}_i};{\overleftarrow{h}_i}\right]$.

Then we feed $h_i$ to an attention-based multi-layer decoder. The parameters of the model are trained to maximize the log-likelihood on the entire training set, which is formulated as:
\begin{equation}
    \mathrm{argmax}\sum\nolimits_{i=1}^M \log P(L'_i|L'_{i-m:i-1}, k_i),
\end{equation}
where M is the number of input-output pairs of the model.

\iffalse
\begin{figure*}[!t]
    \centering
    \includegraphics[scale=0.55]{figs/refinement.png}
    \caption{A graphical illustration of the Refinement stage. When generating line $L_i$, bidirectional context encoder is used to encode context ${L_{i-m:i-1}}$ and ${L}'_{i+1:i+m}$, draft encoder is for $L'_i$. Impressive word candidates consist of keyword $k_i$ and good patterns we detected. ``$bcv$'', ``$dv$'' and ``$ref$'' are bidirectional context vector, draft vector and refining vector, respectively.}
    \label{fig:model}
\end{figure*}
\fi

\subsection{Impressive Words Detection Mechanism}

\begin{algorithm}[H] 
\caption{Impressive Words Detection} 
\label{alg:impressive words detection algorithm} 
%\textbf{Input}: Line $L'$ in draft $D$ (we omit $i$ for simplicity), candidate number $N_{iw}$\\
%\textbf{Output}: Impressive words $W$
\begin{algorithmic}[1]
\REQUIRE 
Line $L'$ (we omit $i$) in draft $D$, candidate number $N_{iw}$
\ENSURE Impressive words $W$
\STATE Retrieve twenty human-written lines $R$ from Elasticsearch based on $L'$ and keyword $k$.
\STATE Segment each line $r \in R$ into words and select one line $r'$ based on Jaccard similarity and sentence length.
\STATE Label POS tags, calculate TFIDF values for each word in $L'$ and $r'$, and keep nouns ($n$.), adjectives ($a$.) and verbs ($v$.).
\STATE Group words: $n$., $a$. for one set and $v$. for the other, and get word lists $wL'_{na}$, $wL'_v$ and $wr'_{na}$, $wr'_v$ for $L'$ and $r'$.
% \FOR{each $item \in \{wl'_{na}, wl'_v, wr'_{na}, wr'_v\}$}
\STATE Sort $wL'_{na}$, $wL'_v$, $wr'_{na}$ and $wr'_v$ individually by TFIDF values in descending order.
% \ENDFOR
\STATE Get new word lists $wL''$, $wr''$ by concatenation, $wL'' = wL'_{na} + wL'_v$, $wr'' = wr'_{na} + wr'_v$.
\STATE Let $cn = 0$, $W = []$
\FOR{each $w \in wr''$}
\IF{$ cn < N_{iw}$}
\IF{$w \notin wL''$}
\STATE Add $w$ to $W$, $cn = cn + 1$
\ENDIF
\ELSE
\STATE Jump out of the loop
\ENDIF
\ENDFOR
\RETURN $W$
\end{algorithmic}
\end{algorithm}

In order to learn good patterns explicitly in poems written by humans and generate new expressions, we import the impressive words detection mechanism.

Given the entire training set, we index each line and construct a query as the combination of the draft line $L'_i$ and its keyword $k_i$. We use the query to retrieve top 20 similar poem lines from the index based on a BM25 score \cite{robertson2009probabilistic}. Here, we use an open-source tool Elasticsearch\footnote{\url{https://www.elastic.co/products/elasticsearch}}. Then, we pick sentences with characters more than 5 to maintain meaningful ones, perform word segmentation by Jieba\footnote{\url{https://github.com/fxsjy/jieba}}. Sentences that are almost identical with the draft line are not needed, since do not want to simply copy the retrieval lines. Our goal is to generate some new and impressive expressions by learning the most essential patterns from retrieval results. Therefore, we pick out the most similar retrieval lines in the range of [0.3, 0.7] (Algorithm \ref{alg:impressive words detection algorithm} line 1 to 2) based on Jaccard similarity which is defined as:

\begin{equation} \label{eq:jaccard_similarity}
    J(S(L'), S(r)) = \frac{|S(L')\cap S(r)|}{|S(L')\cup S(r)|},
\end{equation}

\noindent where $S(L')$ and $S(r)$ are word sets of the draft line $L'$ and retrieval line $r$, respectively. $|\cdot|$ denotes the size of a set.

For the obtained one retrieval line, we employ Part of Speech (POS) tagging on each word by Jieba\footnote{Jieba is for modern Chinese, which fits our task on modern Chinese poetry, and its POS tagging results are reasonable enough by human evaluation.} and only keep nouns($n$.), adjectives($a$.) and verbs($v$.), since they are usually semantically rich. Then we group these three kinds of words into $na$ (nouns and adjectives) set and $v$ (verb) set. For each set, we use the TFIDF value to sort words in descending order. Then we get two concatenated ordered word lists denoted by $wL''$ and $wr''$ (Algorithm \ref{alg:impressive words detection algorithm} line 3 to 6). Then we select words appearing in $wr''$ but not in $wL''$ as the impressive word candidates for line $L'$ (Algorithm \ref{alg:impressive words detection algorithm} line 7 to 17). Finally, we have triples $\{(L'_i, k_i,  \{w_{i,j}\}_{j=1}^{N_{iw}})\}_{i=1}^{n}$ for each draft, where $N_{iw}$ is the number of impressive words candidates.

\subsection{Refinement}
In Refinement stage, we generate a new poem $P$ with $n$ lines, $P = {\{L_i\}_{i=1}^{n}}$ by taking into account both the draft and ``refining vector'' distilled from impressive patterns. When generating line $L_i$, there are three parts of the input, which are bidirectional sentence-level context, $L'_i$ from draft and the ``refining vector''.

%Bidirectional sentence-level context contains both previous lines in $P$, i.e., $L_{<i}$, and also corresponding future lines in $D$, i.e., $L'_{>i}$. 
\subsubsection{Construct Bidirectional Context}
For NLG tasks, when generating a word in a sequence, only previously produced words can be utilized. Even with two decoders like Deliberation Network \cite{xia2017deliberation}, the backward and forward information are limited in one sentence. In contrast, given a draft, humans tend to polish a line based on sentences before and after current one to make the poem more coherent and fluent. Inspired by this, the bidirectional sentence-level context in our model is composed of ${L_{i-m:i-1}}$ and $L'_{i+1:i+m}$. ${L_{i-m:i-1}}$ are lines we generated before $L_i$ in refinement stage and provide information in the past, while $L'_{i+1:i+m}$ are lines in the draft version and represent information in the future. Finally, ${L_{i-m:i-1}}$ and $L'_{i+1:i+m}$ are concatenated and the bidirectional context is transformed to hidden vectors ${\{h_j| h_j = \overrightarrow{h_j} ; \overleftarrow{h_j}\}_{j=1}^{2m}}$ with bidirectional LSTM.

\subsubsection{Refining Vector}
For each draft line $L'_i$, impressive words detection gives out corresponding impressive words candidates $W = \{w_i\}_{i=1}^{N_{iw}}$. Instead of feeding these words and keyword $k_i$ directly to the model, we compute a ``refining vector'' by an attention mechanism defined as follows:

\begin{equation}
    ref = \sum_{w \in W \cup \{k_i\}} \alpha_w\hat{e}_w,
\end{equation}
where $\hat{e}_w$ is the embedding of word $w$. The weight $\alpha_w$ is computed by:
\begin{equation}
    \alpha_{w} = \frac{\exp \left(e_w \right)}{\sum_{w \in W \cup \{k_i\}} \exp \left( e_w \right)},
\end{equation}
\begin{equation} 
    e_w = {\bm{\mathrm{v}}}_\alpha^T
    \tanh\left({\bm{\mathrm{W}}}_\alpha h_{2m} + {\bm{\mathrm{U}}}_\alpha\hat{e}_w\right).
\end{equation}
Here, $h_{2m}$ is the last hidden state of bidirectional context encoder, since we need to consider bidirectional context when we want to add impressive words in current sentence.

The decoder state $s_t$ is updated by:
\begin{equation} \label{eq:decoder_hidden_state}
    s_t = f_{\mathrm{LSTM}} (s_{t-1}, \hat{e}_{y_{t-1}},r_t),
\end{equation} 
%where $r_t$ is defined as:
\begin{equation}
    r_t = bcv_t \oplus dv_t \oplus ref,
\end{equation}
where $bcv_t$ and $dv_t$ denotes the bidirectional context vector and draft vector at time step $t$, $ref$ is time step independent. For bidirectional context and draft hidden states, we apply two attention mechanisms on them, following Equation \ref{eq:att_weigth_sum}. Then we get bidirectional context vector $bcv$ and draft vector $dv$.

\section{Experiments}
\label{sec:experiments}
We first introduce some empirical settings, including the dataset, baselines, implementation details and performance measures, then use evaluations on both automatic metrics and human judgements to prove the effectiveness of our model. Finally, we conduct case study to show the quality of generated poems.

\subsection{Dataset}
\begin{table}[!htb]
    \centering
    \caption{Statistics about our modern Chinese poetry corpus.}
    \begin{tabular}{l c} \toprule[1pt]
         %& Number \\ \midrule[1pt]
         Number of poems in training set & 210,935 \\ 
         Number of poems in validation set & 26,367\\ 
         Number of poems in test set & 26,367\\ 
         Lines per poem & 10.25\\ 
         Characters per line & 12.35\\ 
         Characters per poem & 143.77\\ \bottomrule[1pt]
         %Vocabulary size & 170,342\\ \bottomrule[1pt]
    \end{tabular}
    \label{tab:dataset}
\end{table}

Since there is no public large-scale modern Chinese poetry dataset, we collect a new dataset. Our dataset is constructed with 2 parts: (1) modern Chinese poetry, collected from a online poetry website\footnote{\url{http://www.shigeku.org}}; (2) modern Chinese lyrics, collected from NetEase Cloud Music\footnote{\url{http://music.163.com}}. Lyrics are very close to modern poems in both content and style so we can regard them as poetry. We totally collect 263,669 modern Chinese poems containing 9,209,186 sentences. Then, we tokenize each line to words by Jieba and calculate the TextRank score for each word. The word with the highest TextRank score is selected as the keyword for each line. The dataset is separated into training, validation, and test sets with the ratio 8:1:1. Table \ref{tab:dataset} provides descriptive statistics about our dataset.

\begin{table*}[!htb]
    \centering
    \caption{Automatic and human evaluation results. The last line is the results of human-written poems in test set and human evaluation set. All significance tests are measured by t-test, and the results show that the improvements of our model are significant on the dataset, i.e., p-value $<$ 0.01.}
    \begin{tabular}{l|cccccc|cccc} \toprule[1pt]
        & \multicolumn{6}{ c |}{\bf Automatic Evaluation} 
        & \multicolumn{4}{ c }{\bf Human Evaluation} \\
        \bf & \bf PPL & \bf Rough-L & \bf Distinct-1 & \bf Distinct-2 & \bf Novelty-2 & \bf Novelty-3 & \bf Fluency & \bf Coherence & \bf Impressiveness & \bf Poeticness \\ \midrule[1pt]
        Plan & 30.98 & 0.3375& 0.2978 & 0.7025 & 1020 & 4985 & 3.19 & 2.94 & 2.74 & 2.97 \\ 
        DN & 26.33 & 0.3423 & 0.3017 & 0.7203 & 1036 & 5006 & 3.23 & 3.08 & 2.79 & 3.09 \\ 
        EED & 27.45 & 0.3533& 0.3121 & 0.7655 & 1058 & 5073 & 3.32 & 3.29 & 3.16 & 3.21 \\
        Mem & 27.72 & 0.3425& 0.3335 & 0.7836 & 1077 & 5092 & 3.55 & 3.52 & 3.35 & 3.18 \\ 
        WM & 26.84 & 0.3654& 0.3226 & 0.7521 & 1043 & 5089 & 3.68 & 3.73 & 3.50 & 3.30 \\ 
        \bf GRR-Refine & \bf 24.06 & 0.3178 & 0.3237 & 0.7636 & 1050 & 5070 & 3.85 & 3.94 & 3.44 & 3.37 \\
        \bf GRR & 28.52 & \bf 0.4138 & \bf 0.3447 & \bf 0.8287 & \bf 1085 & \bf 5102 & \bf 3.88 & \bf 3.98 & \bf 3.86 & \bf 3.40 \\ \midrule[1pt]
        Human-written & 25.32 & / & 0.3640 & 0.8275 & 1064 & 5072 & 4.06 & 4.20 & 4.35 & 4.42 \\ \bottomrule[1pt]
    \end{tabular}
    \label{tab:automatic and human eval}
\end{table*}

\subsection{Baselines}
We compare our model with representative poetry generation and refinement approaches as listed below:
%which are Planning-based Model ~\cite{wang2016chinese}, Deliberation Network ~\cite{xia2017deliberation} and Exemplar Encoder-Decoder (EED) ~\cite{pandey2018exemplar}. Since they originally were proposed for traditional Chinese poetry generation, machine translation and dialogue generation separately, we extend each of them to our task as follows:

\textbf{Plan:} a Planning-based model \cite{wang2016chinese} which divides poetry generation into two steps: organizing outlines (keywords) and writing poems.

\textbf{DN:} the Deliberation Network \cite{xia2017deliberation} which is firstly proposed for machine translation. When generating a word in a sentence, it looks backward and forward in the range of current sentence by jointly optimizing two decoders. It's a representative model for the ``generate-then-refine'' paradigm.

% \textbf{2Encoders($C$+$L'$): }We construct bi-directional context $C$ as aforementioned and replace the first encoder input with this context information. 

\textbf{EED:} the Exemplar Encoder-Decoder model \cite{pandey2018exemplar} which is firstly proposed for neural conversation generation. There are two encoders: context encoder and similar-sentence encoder. These similar sentences are retrieved from training set and fed entirely into the second encoder. It's a representative model for the ``retrieval-then-generate'' paradigm.
% Using the draft and keywords, we retrieve similar human written poem lines and directly treat the top1 result given by Elasticsearch as the input. We also add the draft encoder in Figure~\ref{fig:model}. This can be considered as an ablation study to test the validity of IWDIM under the retrieval setting. Sentence-level context here is bidirectional.

\textbf{Mem:} a poetry generation model with neural memory \cite{zhang2017flexible} that contains human-written poems in a static external memory to improve the generated {\it quatrains}. It aims to generate creative Chinese poetry.  

\textbf{WM:} a recent Working Memory model \cite{yi2018chinese} for poetry generation that dynamically invokes a memory component by saving the writing history into memory. It focuses on generating coherent poems.

% \textbf{2Encoders($C$+$L'$)+IWDIM: }Our model utilize two encoders taking bi-directional context and generated draft as inputs respectively. In addition, we use impressive words detection and insertion mechanism to select out good patterns candidates and integrate them into our model.
We denote our model as {\bf GRR}, and {\bf GRR-Refine} is the one without the ``refining vector".

\subsection{Implementation Details}
We employ 54,500 words with the highest frequency as our vocabulary and define all the out-of-vocabulary words to a special token $<$unk$>$. The word embedding size is 128, and initialized with word2vec \cite{Mikolov3:2013} pre-trained on the poetry corpus. The recurrent hidden layers of encoder and decoder contain 128 hidden units, and the number of layers is 4. The model is trained using the Adam algorithm \cite{diederik:2014}, where the batch size is 512 and the learning rate is 3e-4. The dropout technique \cite{srivastava2014dropout} is also adopted and the dropout rate is set to 0.3. The number of sentence-level context in one direction ($m$) is set to 1. The number of impressive word candidates ($N_{iw}$) is set to 2. All models are implemented with the same set of hyper-parameters. Optimization objective is standard cross entropy. For inference, beam search is utilized and the beam size is 10. We tune our hyper-parameters on validation set and measure the performance on test set. We use Tensorflow Framework\footnote{\url{https://www.tensorflow.org/}}
for our implementation.

\subsection{Performance Measures}
%\subsubsection*{Automatic Evaluation Metrics}
We use four metrics for automatic evaluation: \textbf{Perplexity (PPL)}: it measures the average fluency of generated poems. Using a 5-gram character based language model trained on our poetry corpus, we calculate the perplexity on test set. \textbf{Rouge-L}: it uses longest common sub-sequence to calculate the similarity between the generated line and its reference \cite{lin2004rouge}. \textbf{Distinct-1/2:} it reflects whether poems are diverse in content. It is defined as the ratio of unique uni/bi-grams over all uni/bi-grams in generated poems \cite{N16-1014}.
%\textbf{Perplexity (PPL):} it measures the average fluency of generated poems. Using a 5-gram character based language model trained on our poetry corpus, we calculate PPL on test set. The lower PPL is, the more fluent a poem is. Note that a low PPL may also indicate that the sentence consists of some simple words.
%\textbf{Rouge-L:} rough-L uses longest common sub-sequence to calculate the similarity between the generated and human-written poems. A higher rouge score indicates that the generated poem is more similar to its reference. 
\textbf{Novelty-2/3:} it is a new metric defined in this paper, which is calculated as the number of new bi-/tri-grams that do not appear in the training set. 

%Our goal is to generate coherence and novel poems. Therefore, the evaluation metrics based on n-gram overlap such as BLEU and ROUGE are not suitable for our task. Besides, Perplexity (PPL)\footnote{Our method also improves PPL over the baseline methods.} is also not reasonable and not related with coherence.

%\subsubsection*{Human Evaluation Metrics}
Human evaluation is necessary for poetry generation. In order to make our results more believable, we use four criteria for human evaluation following Yi et al. \cite{yi2017generating} and Wang et al. \cite{wang2016chinese}:
\textbf{Fluency:} it measures whether the poem reads smoothly and fluently. 
\textbf{Coherence:} it measures the relevance of adjacent lines in one poem.
\textbf{Impressiveness:} one of our motivations is trying to learn some good patterns explicitly from human-written poems and then to generate new ones. We design this criterion to let annotators judge whether our model generates some impressive expressions. 
\textbf{Poeticness:} it represents the overall quality of a poem, such as whether a poem could convey a poetic image and artistic conception.
%\textbf{Poeticness:} it reflects the \textbf{creativity} of a poem in poetic aspect. 

We randomly select 200 groups of keywords and feed them into 7 models. For each group, we shuffle these 7 poems and the corresponding human-written one, then display them in one page\footnote{Since annotators can see 8 poems at the same time, their scores are based on comparison, which are more reliable.}, and the annotators do not know their sources. During evaluation, annotators can also see retrieval lines to help them judge the novelty to some extent. Each criterion is assessed with a score from 1 (worst) to 5 (best) by 8 annotators, and the average score for each criterion is computed. The annotators are all postgraduate students in literature background, and they took 10 days on average to finish the evaluation. The Fleiss' kappa \cite{fleiss1973equivalence} value is 0.403. 

\begin{figure*}[!t]
    \centering
    \includegraphics[scale=0.65]{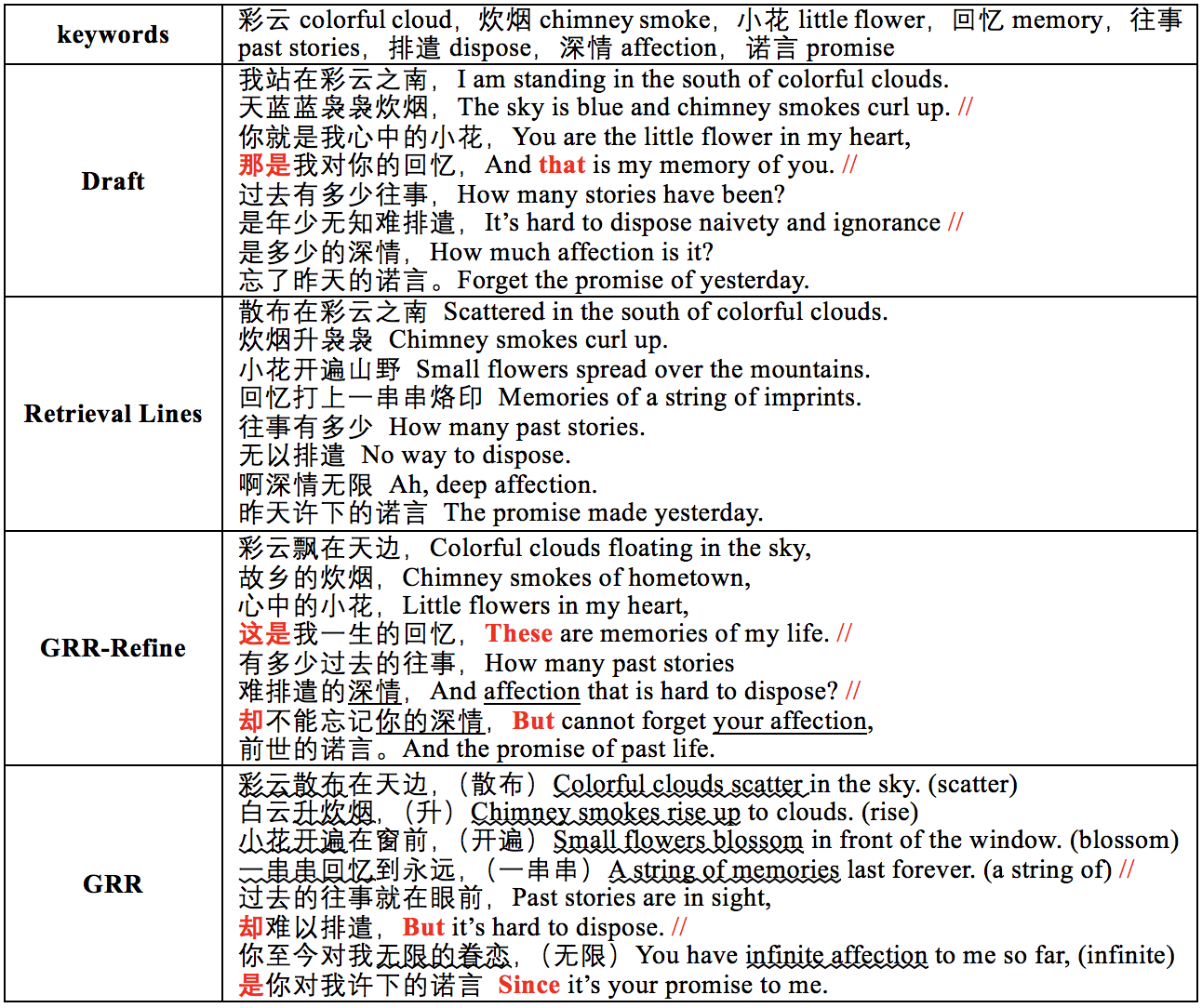}
    \caption{Examples generated by our method. The keywords lie in the first block. The second block is a draft and the third block is corresponding retrieval lines. Words in parentheses in the last block is the impressive words candidates extracted by impressive words detection mechanism. Words in (red) bold play a positive role on coherence, and phrases with wavy underlines are impressive expressions. In order to better understand these generated poems, we also use ``//'' to separate slightly different semantic chunks.}
    \label{fig:casestudy}
\end{figure*}

\subsection{Experimental Results}
Now we demonstrate our experimental results on the dataset in terms of automatic evaluation and human evaluation.

\subsubsection{Automatic Evaluation Results}
The left part of Table \ref{tab:automatic and human eval} shows the automatic evaluation results on our test set. Our proposed method GRR outperforms other models almost on all metrics. GRR-Refine receives the lowest {\it PPL} values, which shows that bidirectional sentence-level context is beneficial to fluency. It has been proven that lower {\it PPL} values usually correspond to simple and general sentences, thus a higher {\it PPL} for GRR indicates that our model can generates more diverse and novel words. A higher {\it Rough-L} score shows that our generated sentences are more similar to their references, which also implies their diversity. 

GRR-Refine or EED help very little on {\it Distinct-1/2} score, even information like future sentences or similar sentences are provided to these models. Compared to EED, Mem and GRR-Refine, the highest {\it Distinct-1/2} score of GRR illustrates the effectiveness of the ``refining vector'', that is, rather than using the entire retrieval candidates, memory-stored sentences or generated lines, the ``refining vector'' is more useful to reduce the noises and improve the diversity. Our approach increases both {\it Distinct-1/2} and {\it Novelty-2/3} significantly, which indicates that it generates more diverse and creative expressions. All significance tests are measured by t-test, and the results show that the improvements of our model are significant on the dataset, i.e., p-value $<$ 0.01. 

\subsubsection{Human Evaluation Results}
Human evaluation results are shown in the right part of Table \ref{tab:automatic and human eval}. GRR receives the best evaluation on all metrics. The evaluation results on {\it Coherence} and {\it Impressiveness} prove that our model can jointly improve coherence, diversity and novelty. % Here, Impressiveness corresponds to Distinct and Novelty in automatic evaluation.
Compared with DN and WM, GRR-Refine has higher {\it Coherence} score, which indicates the validity of bidirectional sentence-level context for improving the coherence. Besides, human evaluation results on {\it Fluency} and {\it Coherence} show that {\it PPL} is not totally reliable for sentence fluency and especially poem coherence. For human, a sentence with diverse language usage can also be fluent but its {\it PPL} value may be high. 

{\it Impressiveness} and {\it Poeticness} scores almost have the same tendency. GRR outperforms other methods significantly, which means that our proposed approach can generate some good and new expression patterns successfully, and more impressive patterns also help express intents and emotions clearly, which contributes to the overall {\it Poeticness}. All significance tests are measured by t-test, and the results show that the improvements of our model are significant on the dataset, i.e., p-value $<$ 0.01. 

Since human-written poems are mixed with other generated ones and evaluated by annotators, we put the results of them in the last line of Table \ref{tab:automatic and human eval} to show the gap between generated and human-written poems in our human evaluation set.

\subsection{Case Study}
We present examples generated by GRR-Refine and GRR with the same keywords in Figure \ref{fig:casestudy} for case study. Comparing the draft and the poem generated by GRR-Refine, we figure out that the correct usage of conjunctions and pronouns can improve the coherence of entire poem. For the draft, ``you are the little flower in my heart, and that is my memory of you'', this sentence is about the memory of a person and does not have many connections with preceding and following lines. In contrast, the fourth line in GRR-Refine poem writes that ``these are memories of my life'', and looking back to the first three lines in it, they are all in the structure that a noun with its modifier, which enhances the relationship in the first four lines. Word ``affection'' with underlines in the sixth line is also the keyword in the seventh line, which shows that the generation of the sixth line is influenced by its next line. Above characteristics also occur in the poem generated by GRR, for example, ``but'' and ``since'' are well used in this example. The reason for improving coherence is that we take bidirectional context into consideration, and this can help generate closely tied sentences.

When it comes to impressive words generation, phrases with wavy underlines in GRR poem are new patterns that do not appear in neither the draft nor corresponding retrieval lines. A poem can be creative and vivid if it includes various nouns, verbs and adjectives. As we can see, ``scatter'', ``blossom'' and ``a string of'' are all impressive expressions that can lighten a sentence. The GRR poem shows that our generated poems can be more diverse and creative.

\section{Conclusion and Future Work}
\label{sec:conclusion}
In this paper, we propose a generate-retrieve-then-refine paradigm for poetry generation by imitating humans' composition process. It enables a generative model to leverage both generated draft and retrieval results. To improve the coherence, we use bidirectional sentence-level context from previous generated lines and draft lines. Also, we introduce the ``refining vector'' distilled by impressive words detection mechanism to generate newer and more impressive expressions. Experimental results on a large-scale modern Chinese poetry dataset show that our model outperforms baselines in terms of coherence and novelty. In the future, we will use other datasets to demonstrate the effectiveness of our approach, and further investigate the way to fulfill impressive words detection in an End-to-End framework.

\section*{Acknowledgements}
We sincerely thank the anonymous reviewers for their helpful and valuable suggestions.

\bibliographystyle{IEEEtran}
\bibliography{IJCNN2020}

\end{document}